\documentclass[10pt,twocolumn,letterpaper]{article}

\usepackage{iccv}
\usepackage[accsupp]{axessibility}
\usepackage{times}
\usepackage{epsfig}
\usepackage{graphicx}
\usepackage{amsmath,bm}
\usepackage{amssymb}
\usepackage{tikz}
\usepackage{multirow}
\usepackage{makecell}
\usepackage{booktabs}
\usepackage{arydshln} 
\usepackage{cite}
\usepackage{wrapfig}

\usepackage[pagebackref=true,breaklinks=true,letterpaper=true,colorlinks,bookmarks=false]{hyperref}

\iccvfinalcopy % *** Uncomment this line for the final submission

\def\iccvPaperID{7} % *** Enter the ICCV Paper ID here

\ificcvfinal\fi

\DeclareMathOperator*{\argmin}{arg\,min}

\begin{document}

%%%%%%%%% TITLE
\title{Personalized 3D Human Pose and Shape Refinement}

\author{Tom Wehrbein\textsuperscript{1,2}\footnotemark[2]
% For a paper whose authors are all at the same institution,
% omit the following lines up until the closing ``}''.
% Additional authors and addresses can be added with ``\and'',
% just like the second author.
% To save space, use either the email address or home page, not both
\and
Bodo Rosenhahn\textsuperscript{1}
\and
Iain Matthews\textsuperscript{2}
\and
Carsten Stoll\textsuperscript{2}
\and
\textsuperscript{1}Leibniz University Hannover, \textsuperscript{2}Epic Games\\
{\tt\small wehrbein@tnt.uni-hannover.de}
%Second Author\\
%Institution2\\
%First line of institution2 address\\
%{\tt\small secondauthor@i2.org}
}

\maketitle
% Remove page # from the first page of camera-ready.
\ificcvfinal\thispagestyle{empty}\fi

%{
%   \renewcommand{\thefootnote}%
%    {\fnsymbol{0}}
%  \footnotetext[1]{$^\dagger$Work primarily done during an internship at Epic Games.}
%}

%%%%%%%%% ABSTRACT
\begin{abstract}
\renewcommand{\thefootnote}{\fnsymbol{footnote}}
\footnotetext[2]{Work primarily done during an internship at Epic Games.}
Recently, regression-based methods have dominated the field of 3D human pose and shape estimation.
Despite their promising results, a common issue is the misalignment between predictions and image observations, often caused by minor joint rotation errors that accumulate along the kinematic chain.
To address this issue, we propose to construct dense correspondences between initial human model estimates and the corresponding images that can be used to refine the initial predictions.
To this end, we utilize renderings of the 3D models to predict per-pixel 2D displacements between the synthetic renderings and the RGB images.
This allows us to effectively integrate and exploit appearance information of the persons.
Our per-pixel displacements can be efficiently transformed to per-visible-vertex displacements and then used for 3D model refinement by minimizing a reprojection loss.
To demonstrate the effectiveness of our approach, we refine the initial 3D human mesh predictions of multiple models using different refinement procedures on 3DPW and RICH. 
We show that our approach not only consistently leads to better image-model alignment, but also to improved 3D accuracy.
\end{abstract}

%%%%%%%%% BODY TEXT
\section{Introduction}
\begin{figure}
\begin{center}
\includegraphics[width=0.9\linewidth]{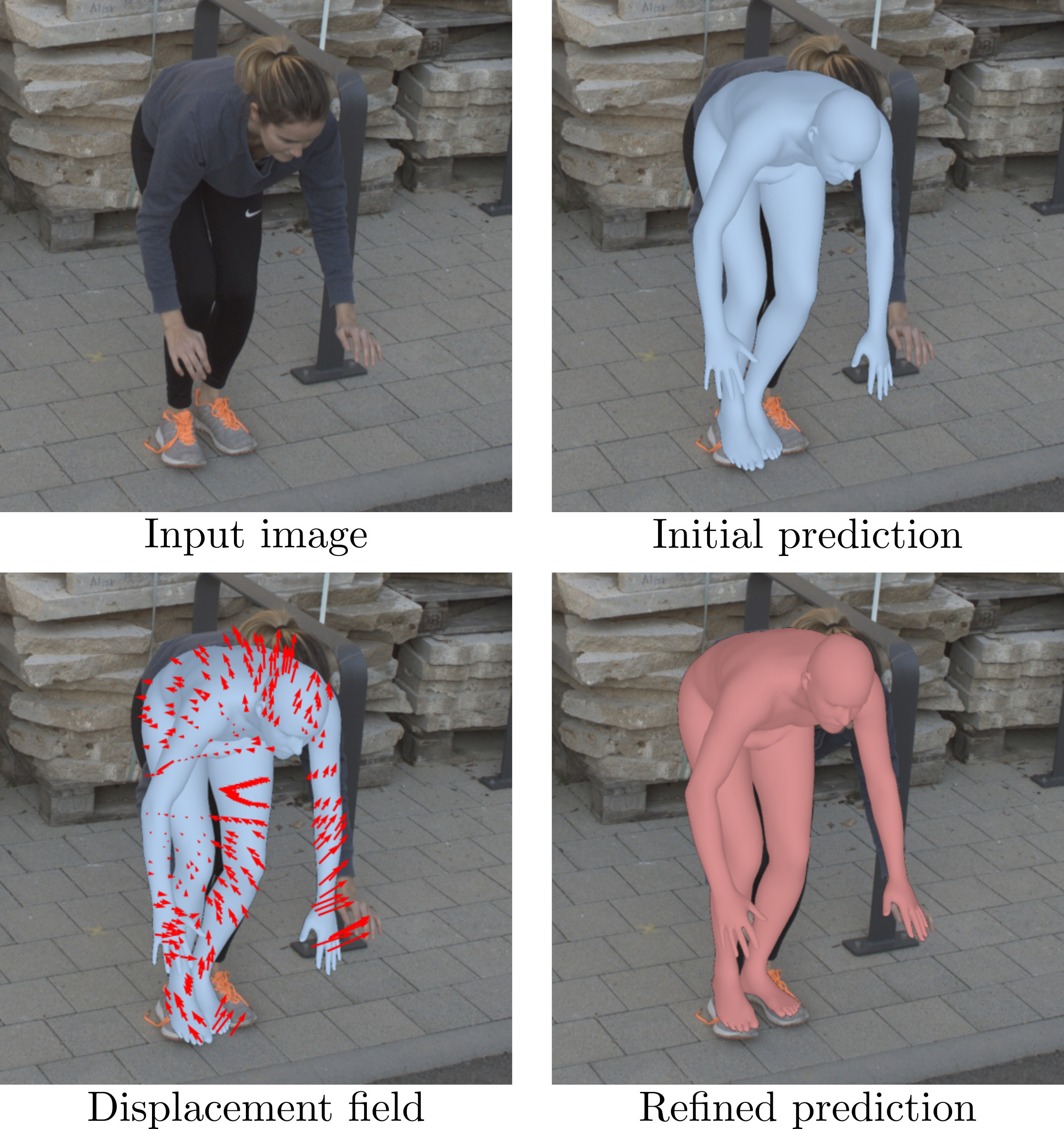}
\caption{Given an initial 3D human model estimate, we predict per-pixel 2D displacements between renderings of the 3D model and the given image that we subsequently use to refine the initial prediction. For clarity, only a sparse subset of displacement vectors is shown. Image is taken from RICH~\cite{huang2022rich}.}
\label{fig:teaser}
\vspace{-1.5em}
\end{center}
\end{figure}

Reconstructing 3D human pose and shape from RGB images is a long-standing and fundamental problem in computer vision due to its various applications in \eg medicine, sports, AR/VR and animation.
Powered by deep CNNs and vision transformers, regression-based methods have made rapid progress and achieve state-of-the-art performance. 
Given a single image or video sequence, they learn to predict the parameters of a human body model (\eg SMPL~\cite{loper2015smpl}) in a data-driven way. 
Despite the promising results and high efficiency, regression-based methods typically suffer from coarse alignment between predicted meshes and image evidence~\cite{zhang2021pymaf} (see Fig.~\ref{fig:teaser}, \textit{top right}).
This is often caused by minor joint rotation errors that accumulate along the kinematic chain, resulting in noticeable drift of joint positions.
Furthermore, the non-linear mapping from image features to global body model parameters together with the complex nature of human appearances makes human body representations extremely difficult to regress accurately without any form of error-feedback loop.
Nevertheless, high-precision estimates are crucial in various applications, especially when interacting with other people or objects in the (virtual) world.

Recently, methods have been introduced that propose to refine an initial regressed human mesh prediction \cite{kolotouros2019spin,joo2020eft,li2022cliff,syguan2021boa,kolotouros2021prohmr,tiwari2022posendf,zhang2021pymaf,guler2019CVPR}.
They focus on either generating 3D pseudo-annotations \cite{kolotouros2019spin,joo2020eft,li2022cliff}, adapting a model to out-of-domain videos \cite{syguan2021boa}, or on the more general task of improving the 3D accuracy for unconstrained monocular images \cite{kolotouros2021prohmr,tiwari2022posendf,zhang2021pymaf,guler2019CVPR}.
All of these methods primarily rely on a data term defined as the reprojection loss between given 2D joints and the projection of the regressed body model joints. 
Target 2D joints are either obtained by an off-the-shelf 2D pose estimator such as OpenPose~\cite{cao2019openpose}, or manually annotated in an offline setting.
However, the resulting human meshes are extremely sensitive to the quality of the given 2D joints.
Joo \etal~\cite{joo2020eft} observed that even manually annotated keypoints often contain non-negligible errors, causing artifacts such as foreshortening in 3D.
They completely ignore the hip and ankle keypoints, since they found them to be particularly noisy.
This sensitivity is even more pronounced when using 2D joint predictions, making it extremely challenging to achieve improvement of the initial 3D body estimate.
There are a lot of cases in which the post-processing step even leads to a degradation in 3D accuracy \cite{joo2020eft,kolotouros2021prohmr,tiwari2022posendf,guler2019CVPR}.
We introduce a drop-in replacement for sparse 2D keypoints that can be used for refining 3D human mesh predictions without requiring manual annotations, and show that the typical 25 keypoints used by \cite{cao2019openpose} are not sufficient to robustly constrain the full human body.

Instead of using keypoints for refinement, we learn dense 2D correspondences that can be effectively used as image cues for refining estimated 3D human meshes in realistic and challenging in-the-wild scenarios.
We leverage initial 3D mesh estimates generated by state-of-the-art regression-based pose estimators~\cite{kolotouros2021prohmr, kocabas2021pare, pang2022hmr+} and learn 2D displacements between renderings of the predictions and the corresponding images.
This allows us to integrate and exploit appearance information of the person and utilize 3D information in the form of depth and normal renderings.
By taking into account the initial human mesh prediction, the network only has to learn small displacements while being able to adapt to typical prediction errors.
Furthermore, only image displacements need to be learned compared to complex pixel to 3D body surface mappings required by DensePose~\cite{Guler2018DensePose,guler2019CVPR}.
Instead of designing a specialized regressor architecture to improve image-model alignment~\cite{zhang2021pymaf}, our approach can be combined with any 3D human mesh regressor and benefits from advances in that field, as well as advances in optimization techniques \cite{ChoutasECCV2022,song2020eccv}.
Using 2D correspondences for refinement leads to better image-model alignment and to improved 3D accuracy.
As shown in Fig.~\ref{fig:teaser}, even accurate 3D estimates can be further refined.

To evaluate our approach, we refine the initial 3D human mesh predictions of multiple models \cite{kocabas2021pare,kolotouros2021prohmr,pang2022hmr+} using different fitting procedures \cite{Bogo2016ECCV,kolotouros2021prohmr,pavlakos2019smplx} on 3DPW~\cite{marcard2018eccv} and RICH~\cite{huang2022rich}.
We compare the performance of using OpenPose~\cite{cao2019openpose}, DensePose~\cite{Guler2018DensePose} and our displacement fields for the reprojection loss in optimization.
We show that our displacement fields lead to significantly better performance in almost all settings and metrics.
Additionally, we demonstrate that our model is robust to noisy and erroneous texture estimates, as well as to changes in illumination.

To summarize, our main contributions are:
\begin{itemize}
    \item We present a method to learn per-pixel 2D correspondences between renderings of a 3D human mesh and an image that enables 3D human mesh refinement. 
    \item The appearance information of the person is successfully leveraged to boost prediction accuracy.
    \item Our 2D displacement fields can refine the estimates of off-the-shelf 3D human mesh regressors and consistently outperform OpenPose keypoints for refinement.
\end{itemize}

\section{Related Work}
The de facto approach for monocular 3D human mesh recovery is to estimate the low-dimensional parameters of a statistical body model~\cite{anguelov2005scape,loper2015smpl,pavlakos2019smplx,xu2020ghum} such as SMPL~\cite{loper2015smpl}.

\textbf{Optimization-based approaches} have historically been the leading paradigm for model-based 3D human mesh estimation.
They rely on classical optimization to iteratively fit the body model parameters to 2D image observations.
Pioneering work in this area \cite{sigal2007nips,guan2009iccv,hasler2010cvpr} leveraged 2D keypoints or silhouettes for human body fitting but required manual user intervention.
Enabled by advances in 2D human pose estimation \cite{pishchulin2016cvpr}, Bogo \etal~\cite{Bogo2016ECCV} introduced SMPLify, the first fully automated approach.
SMPLify fits the SMPL model to detected 2D keypoints utilizing multiple strong priors to regularize the optimization. 
Subsequent work investigated different data terms, \eg silhouettes \cite{lassner2017UP}, part orientation fields (POFs) \cite{xiang2019monocular}, dense correspondences \cite{guler2019CVPR} and contact information \cite{Mueller2021cvpr,taheri2020grab}, extended the approach to multi-view and multi-person \cite{zanfir2018fitting,dong2021shape,huang2017mvsmplify}, and devised more efficient optimization pipelines \cite{fan2021revitalizing}.
However, due to their robustness and performance on challenging in-the-wild data, recent methods \cite{kolotouros2021prohmr,pavlakos2019smplx,joo2020eft,li2022cliff,kolotouros2019spin} almost exclusively rely on 2D skeletons estimated by off-the-shelf pose estimators \cite{cao2019openpose}.
To better constrain the 3D body during fitting and thus reduce ambiguities, recent work focused on constructing stronger 3D pose priors \cite{kolotouros2021prohmr,pavlakos2019smplx,davydov2022gans,tiwari2022posendf} or on training neural optimizers \cite{ChoutasECCV2022,song2020eccv,Zanfir2021CVPR} to predict the parameter updates.
In general, optimization-based approaches achieve well-aligned results, but tend to be sensitive to initialization and the quality of the given image cues.

\textbf{Regression-based approaches} \cite{kanazawa2018hmr,kolotouros2019spin,kocabas2021pare,kolotouros2021prohmr,li2022cliff,straps2020bmvc,Zhang2020CVPR,li2021hybrik,kocabas2021spec,wei2022mpsnet,li2023niki} use a deep network to predict 3D body parameters directly.
To compensate for the lack of in-the-wild 3D annotations, methods have focused on integrating alternative supervision signals.
They often rely on 2D annotations, such as keypoints \cite{kanazawa2018hmr,vince2017bmvc}, silhouettes \cite{pavlakos2018cvpr,straps2020bmvc,tung2017nips}, part segmentations \cite{zanfir2020weaklyNF,kocabas2021pare,Dwivedi2021iccv,omran3DV2018}, or dense correspondences \cite{zeng2020cvpr,Zhang2020CVPR,xu2019denserac,guler2019CVPR,zhang2020densepose2smpl}, that can be integrated as reprojection losses or leveraged as proxy representations.
Regression-based approaches are fast and achieve state-of-the-art reconstruction performance.
However, since they lack an error-feedback loop, they typically suffer from coarse alignment between predicted meshes and image evidence \cite{zhang2021pymaf}.
Recently, aiming at producing well-aligned meshes, Zhang~\etal~\cite{zhang2021pymaf} introduced PyMAF, a specialized deep regressor with an integrated alignment feedback loop that leverages learned feature pyramids.

\textbf{Hybrid approaches.}
To combine the best of both paradigms, recent work has explored hybrid approaches.
\cite{pavlakos2018cvpr} demonstrated that by initializing \mbox{SMPLify} with their regressed pose parameters the fitting procedure is three times faster and converges to better solutions than vanilla SMPLify.
SPIN \cite{kolotouros2019spin} also uses a regression network to initialize the optimization and leverages the fitted estimates to supervise the network.
This approach has been extended to multi-view by \cite{li2021wacv}.
With the goal of generating 3D pseudo-annotations for 2D datasets, EFT~\cite{joo2020eft} updates the network weights for each frame to achieve better reprojection accuracy.  
In a similar manner, BOA~\cite{syguan2021boa} adapts a trained network to out-of-domain streaming videos.
All of the above methods exclusively rely on sparse 2D keypoints as image evidences.
HoloPose~\cite{guler2019CVPR} introduces a refinement procedure that penalizes deviations between the regressed body model and DensePose/2D/3D joint predictions.
However, while the image alignment improves, the 3D accuracy slightly degrades when using DensePose and/or 2D joints for refinement.
Similar observations have been made by \cite{joo2020eft,kolotouros2021prohmr,tiwari2022posendf}, emphasizing the difficulty of fitting 3D model estimates to image cues, especially if the image cues are sparse and noisy and the model estimates are already good.

\section{Method}
\begin{figure*}
\begin{center}
\includegraphics[width=0.9\linewidth]{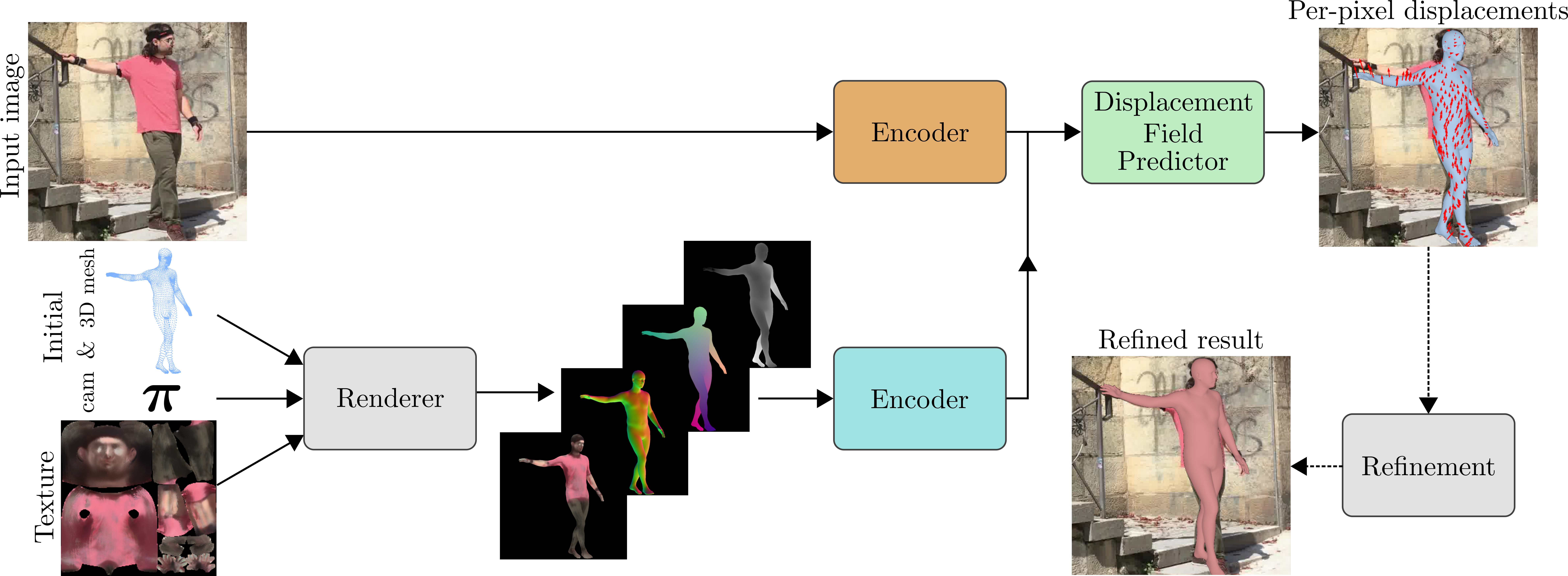}
\caption{Overview of our proposed approach. Given an image with estimated 3D human mesh, camera parameters $\bm{\mathrm{\pi}}$ and an approximate texture map of the target person, we predict the per-pixel 2D displacement field between the 3D human model renderings and the image. The per-pixel 2D displacements are transformed to per-visible-vertex displacements and can subsequently be used to refine the 3D human model using \eg SMPLify~\cite{Bogo2016ECCV}. For clarity, only a sparse subset of displacement vectors is shown. Reference image from 3DPW~\cite{marcard2018eccv}.
}
\label{fig:method}
\vspace{-1.5em}
\end{center}
\end{figure*}

Our aim is to construct dense correspondences between an initial human mesh prediction and the given image that can be used to refine the initial prediction and thus improve the accuracy of the 3D mesh.
Motivated by progress and applications in 3D human texture estimation \cite{xu2021texformer,rajasegaran2021tracking,iqbal2022rana,alldieck2018video,alldieck19cvpr,pavlakos2019texturepose}, we aim to exploit the appearance of the person for 3D mesh refinement.
Given a short calibration sequence where the person is seen from all sides, an accurate texture map can be computed \cite{alldieck2018video,alldieck19cvpr,iqbal2022rana}.
If no calibration sequence is available, a rough texture map can be build over time \cite{rajasegaran2021tracking}.
We argue that for almost all practical applications, first estimating the texture map of the person is non-intrusive and adds very little overhead.
Inspired by the 6D object pose estimation refinement approach of Grabner \etal~\cite{grabner2020eccv}, we learn pixel-wise 2D displacement fields between the 3D human model renderings and the images similar to optical flow \cite{teed2020raft,xu2022gmflow,huang2022flowformer,fischer2015cvpr}.
Unlike Grabner \etal~\cite{grabner2020eccv}, we utilize the estimated texture maps to generate RGB renderings.
Additionally, we use depth, normal and unique vertex color renderings to explicitly provide 3D information.
We train a CNN-based network (Sec.~\ref{model}) that takes as input the model renderings together with the image and outputs a 2D displacement vector for each rendered pixel.
The per-pixel displacements can then be efficiently transformed to per-visible-vertex displacements utilizing information provided by the renderer.
Finally, we use the per-vertex vectors to perform 3D model refinement (Sec.~\ref{refinement}).
This way, an ideal geometric reprojection loss of the full human mesh can be minimized.
The overall framework is depicted in Fig.~\ref{fig:method}.

\textbf{Body representation.}
We use SMPL~\cite{loper2015smpl} to represent the human body. It provides a differentiable function that given pose $\bm{\theta} \in \mathbb{R}^{72}$ and shape $\bm{\beta} \in \mathbb{R}^{10}$ parameters outputs a 3D mesh $\mathcal{M}(\boldsymbol{\theta}, \boldsymbol{\beta}) \in \mathbb{R}^{N\times3}$ with $N=6890$ vertices.
In addition, the 3D body joints $\mathcal{J}_\mathit{3D}$ can be expressed as a linear combination of the mesh vertices.
A linear regressor $W$ can be pretrained for this task to produce the $k$ joints of interest $\mathcal{J}_\mathit{3D}=W \mathcal{M} \in \mathbb{R}^{k\times3}$.

\subsection{Model Design} \label{model}
Given a single RGB image, an initial 3D human mesh and camera estimate and an approximate texture map of the person, we compute per-pixel 2D displacements between the 3D human model renderings and the image.
Formally, the per-pixel displacement field $\mathbf{f}: \mathbb{N}^{2} \rightarrow \mathbb{R}^2$ maps every valid 2D pixel location $\mathbf{x} \in \mathbb{N}^{2}$ of the renderings $\mathbf{I}_r$ to its corresponding 2D location $\mathbf{p} = \mathbf{x} + \mathbf{f}(\mathbf{x})$ of the  target RGB image $\mathbf{I}_t$.
A 2D pixel position is considered valid if a pixel has been rendered at that position,
therefore correspondences are not learned for background pixels.

The first step in our pipeline is to render the initial 3D human mesh using the estimated camera parameters and texture map.
In addition to RGB, we generate depth, normal and unique vertex color renderings.
Thus, important 3D information is provided to the network.
The unique per-vertex color attributes are defined as the 3D vertex positions of the neutral SMPL body with mean shape and pose parameters.
Given the output of the rasterizer and the 3D model, the unique vertex color rendering is computed by interpolating the vertex color attributes.
The different renderings are concatenated along the channel dimension.
Next, in order to predict 2D displacements we map the renderings and the input RGB image to a common feature space.
We utilize two different feature encoder branches for this task.
The architecture of both branches is similar to the first stage of ResNet-50~\cite{he2016resnet}.
The only adjustment we make is to use a stride of 1 for all convolutional layers and to remove all max pooling layers.
This maintains the input image size which makes it easier to predict fine-grained displacements.
The architecture of the input image and the renderings feature branches only differ in the number of input channels.
After mapping both input modalities to the common feature space, we concatenate the feature maps and use a stacked hourglass network~\cite{newell2016hourglass} with 4 stacks to predict the 2D per-pixel displacement field.
We train all network branches end-to-end from scratch.
By explicitly predicting the displacement fields, the network learns to be robust to noisy texture maps and changes in illumination.
Additionally, the absence of the scene background in the renderings can be easily dealt with. 
We found that despite the similarity in task, deep optical flow models (\eg \cite{teed2020raft,xu2022gmflow,huang2022flowformer}) do not perform well, even when retrained on human mesh data and when two separate feature encoders are used.
We hypothesize that the 4D correlation volume build by these methods cannot effectively handle large differences in illumination between the rendered and input image.
Furthermore, it is extremely challenging to learn a feature mapping that makes correlating normal and depth features with image features meaningful.

\textbf{Optimization.}
We train our model in a fully-supervised manner.
Given an image with corresponding ground-truth SMPL pose $\hat{\bm{\theta}}$ and shape $\hat{\bm{\beta}}$ parameters and camera projection function $\hat{\bm{\mathrm{\pi}}} : \mathbb{R}^{3} \rightarrow \mathbb{R}^2$, together with a second set of SMPL and camera parameters ${\bm{\theta}}$, ${\bm{\beta}}$,  $\bm{\mathrm{\pi}}$, we first obtain the ground-truth 2D per-vertex displacement field $\hat{\mathbf{v}} \in \mathbb{R}^{N\times2}$ between the projection of both 3D human meshes:
\begin{equation}
\hat{\mathbf{v}} = \hat{\bm{\mathrm{\pi}}}(\mathcal{M}(\hat{\bm{\theta}},\hat{\bm{\beta}})) - \bm{\mathrm{\pi}}(\mathcal{M}(\bm{\theta},\bm{\beta})).
\end{equation}
The parameters ${\bm{\theta}}$, ${\bm{\beta}}$, $\bm{\mathrm{\pi}}$ are either obtained by random perturbations of the corresponding ground-truth parameters or by using the regressed values of some pretrained SMPL prediction model (\eg \cite{kolotouros2021prohmr,kocabas2021pare,pang2022hmr+}).
Since we supervise on the pixel level, the per-vertex displacement field needs to be transformed to a per-pixel displacement field.
This is done by interpolating the per-vertex displacements across the projected triangle surfaces using barycentric coordinates.
Formally, the 2D ground-truth displacement of the pixel at position $(x, y)$ is computed as:
 \begin{equation}
 \label{eq:verts_to_pixel}
\hat{\mathbf{f}}_{x,y} = \sum_{i=1}^{3} b_{x,y,i} \cdot \hat{\mathbf{v}}_{\bigtriangleup_{\texttt{IndexMap}_{x,y,i}}}
\end{equation}
where $\bigtriangleup_{\texttt{IndexMap}_{x,y,i}}$ indexes the $i$-th vertex of the triangle visible at $(x, y)$ and $b_{x,y,i}$ is the corresponding barycentric coordinate.

Finally, we supervise our network on the $l_1$ distance between the ground-truth and predicted 2D per-pixel displacement field:
\begin{equation}
    \mathcal{L} = \frac{1}{WH} \sum_{x=1}^{W}\sum_{y=1}^{H} m_{x,y} ||\hat{\mathbf{f}}_{x,y} - \mathbf{f}_{x,y}||_1, 
\end{equation}
where $m_{x,y}$ is $1$ if a pixel has been rendered at position $(x, y)$ and $0$ otherwise.
The loss is applied at the end of each stacked hourglass stack and the output of the last layer is used as the final prediction.

\subsection{3D Human Mesh Refinement} \label{refinement}
SMPLify~\cite{Bogo2016ECCV} is a popular optimization-based method that fits the SMPL body model to a set of sparse 2D keypoints.
The objective function it minimizes consists of a re-projection term encouraging the 3D body model to explain the observed 2D keypoints and of pose and shape priors that regularize the fit.
More specifically, the optimal fit is given by:
\begin{equation}
\label{eq:smplify}
 \begin{aligned}
(\bm{\theta}^*, \bm{\beta}^*, \bm{\mathrm{\pi}}^*) =  \argmin_{\bm{\theta},\bm{\beta},\bm{\mathrm{\pi}}} &\lambda_{\mathit{2D}}\mathcal{L}_\mathit{2D} + \\
& \lambda_{\bm{\theta}}\mathcal{L}_{\bm{\theta}} + \lambda_{\bm{\beta}}\mathcal{L}_{\bm{\beta}} + \lambda_{\alpha}\mathcal{L}_{\alpha},
 \end{aligned}
\end{equation}
with re-projection term $\mathcal{L}_\mathit{2D}$, 3D pose prior $\mathcal{L}_{\bm{\theta}}$, shape regularizer $\mathcal{L}_{\bm{\beta}}$ and joint bending term $\mathcal{L}_{\alpha}$.
The re-projection term in the original paper calculates the distance between estimated 2D pose keypoints such as~\cite{cao2019openpose} and the corresponding projected joint locations of the SMPL model.
Since the camera parameters are usually unknown, they must be optimized together with the body parameters.
The bending term $\mathcal{L}_{\alpha} = \sum_{i \in (\mathrm{elbows,knees})}\exp(\bm{\theta}_i)$ penalizes unnatural rotations of elbows and knees, the shape regularizer is given as $\mathcal{L}_{\bm{\beta}} = \| \bm{\beta} \|^2$ and the 3D pose prior $\mathcal{L}_{\bm{\theta}}$ is expressed via a Gaussian mixture model.
In order to better constrain the 3D body during fitting and thus reduce ambiguities, recent work~\cite{pavlakos2019smplx,kolotouros2021prohmr,davydov2022gans,tiwari2022posendf} focused on designing stronger 3D pose priors to replace the GMM.
However, the success of fitting the parametric body model depends heavily on the initialization, the balance of data and prior terms and the quality of the sparse 2D keypoints~\cite{joo2020eft,Bogo2016ECCV,lassner2017UP}.

We approach the problem from a different angle and argue that a main source of ambiguity is the insufficient data term $\mathcal{L}_\mathit{2D}$.
There can be a lot of different mesh configurations that explain the observed sparse 2D keypoints~\cite{wehrbein2021iccv,kolotouros2021prohmr}.
Motivated by these limitations, we propose to replace the sparse 2D keypoints with our dense per-pixel displacement fields.
To do this, the predicted per-pixel 2D displacement vectors need to be transformed to per-vertex displacements.
For vertex $i$, this is achieved by accumulating the 2D vectors for all pixel positions $(x, y)$ for which the vertex $i$ is a vertex of the triangle visible at $(x, y)$.
Formally, the displacement $\mathbf{v}_i$ of vertex $i$ is computed as:
\begin{equation}
 \begin{aligned}
\mathbf{v}_i = &\frac{1}{\sum_{x,y} b_{x,y,i}} \sum_{x,y} \left( b_{x,y,i} \cdot \mathbf{f}_{x,y} \right) \\
&\forall x, y : i \in \bigtriangleup_{\texttt{IndexMap}_{x,y}},
 \end{aligned}
\end{equation}
where $\mathbf{f}_{x,y}$ is the predicted 2D displacement at pixel position $(x, y)$ and $b_{x,y,i}$ is the barycentric coordinate of vertex $i$ at that position\footnote{Note that we slightly abuse the notation of Eq.~\ref{eq:verts_to_pixel} and index the barycentric coordinate with the mesh vertex index.}.
The indices of the vertices of the triangle visible at $(x, y)$ are given by $\bigtriangleup_{\texttt{IndexMap}_{x,y}}$.
Finally, we obtain the target 2D vertices $\mathcal{V}_{\mathrm{est}} \in \mathbb{R}^{N\times2}$ by simply adding the predicted displacement field and the projection of the initial model parameters:
\begin{equation}
\mathcal{V}_{\mathrm{est}} = \mathbf{v} + \tilde{\bm{\mathrm{\pi}}}(\mathcal{M}(\tilde{\bm{\theta}},\tilde{\bm{\beta}})).
\end{equation}
Instead of using distance between 2D joint locations we then define the re-projection term of Eq.~\ref{eq:smplify} as: 
\begin{equation}
\label{eq:2d_verts_reproj}
\mathcal{L}_\mathit{2D} =  \sum_{i \in \mathrm{vertices}} w_i \rho(\bm{\mathrm{\pi}}(\mathcal{M}(\bm{\theta},\bm{\beta}))_i - \mathcal{V}_{\mathrm{est}, i}),
\end{equation}
where $w_i$ equals $1$ if vertex $i$ is visible in the rendering and $0$ otherwise, and $\rho$ represents a robust Geman-McClure error~\cite{Geman1987StatisticalMF}.
In addition to providing significantly more 2D landmarks that better constrain the human body during fitting, optimization is no longer dependent on the potentially slightly inaccurate linear mesh-to-joint regressor $W$.
In the experimental section, we show that using our estimated dense 2D displacement, we are able to consistently improve the fitting results over sparse landmark approaches.

\begin{figure}
\begin{center}
\includegraphics[width=1.0\linewidth]{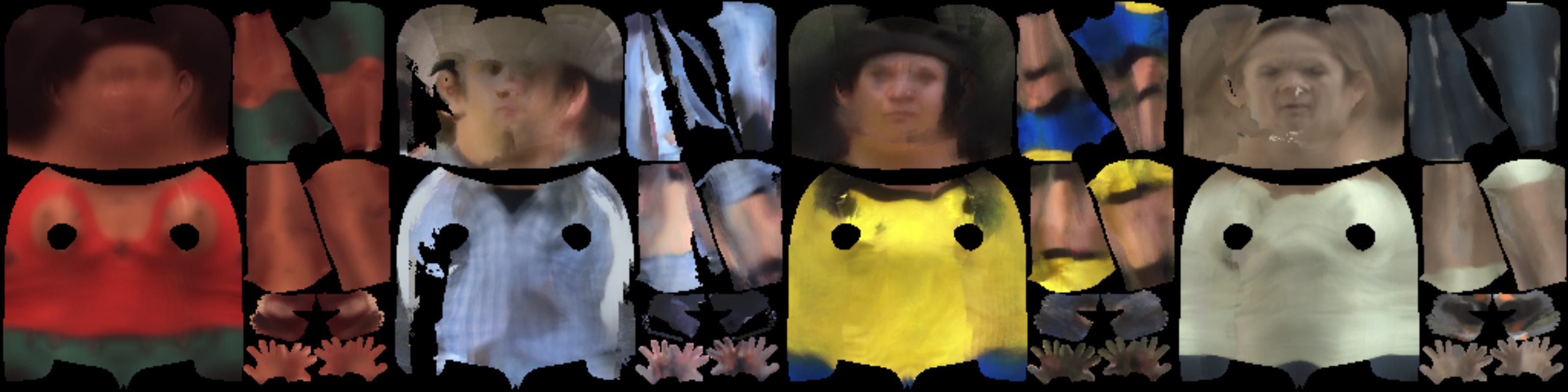}
\caption{Examples of reconstructed texture maps from Human3.6M, 3DPW and RICH used for training and evaluation. We generate texture maps by back-projecting the image colors from multiple frames to all visible vertices.}
\label{fig:texture}
\vspace{-1.5em}
\end{center}
\end{figure}

\section{Experiments}
\textbf{Training.}
We train our model on the standard training sets of Human3.6M~\cite{ionescu2014h36m}, 3DPW~\cite{marcard2018eccv} and SURREAL~\cite{varol2017surreal} using ground-truth camera and SMPL annotations.
Since we want to learn 2D displacements between a rendered 3D mesh and the person in the image, a rich set of SMPL and camera parameter predictions per training image is required.
For every training image, we precompute SMPL and camera parameter predictions using PARE~\cite{kocabas2021pare} and ProHMR~\cite{kolotouros2021prohmr}.
We utilize the probabilistic characteristic of ProHMR and sample 64 predictions for each frame.
To focus on fine-grained displacements, we additionally use ground-truth pose with PARE predicted shape and camera parameters during training.
We perform the rendering on-the-fly at the start of each iteration using nvdiffrast~\cite{laine2020diffrast}.
Since we do not need to keep track of gradients, rendering a batch of 8 only takes around $1$\,ms and thus causes almost no overhead.
For further implementation and training details, we refer the reader to the supplemental material.

\begin{table*}% [t]\scriptsize
	\centering
 \resizebox{\linewidth}{!}{
	\begin{tabular}{lcccccc|cccccc}
		\toprule
	       \multicolumn{8}{c}{3DPW}  & \multicolumn{4}{c}{RICH} \\
		\cmidrule(lr){2-7}\cmidrule(lr){8-13}
		Method + SMPLify   &
		\multicolumn{1}{c}{MPJPE $\downarrow$} & \multicolumn{1}{c}{PA-MPJPE $\downarrow$} & \multicolumn{1}{c}{N-MPJPE $\downarrow$} & \multicolumn{1}{c}{PVE $\downarrow$} & \multicolumn{1}{c}{PA-PVE $\downarrow$} & \multicolumn{1}{c|}{N-PVE $\downarrow$} & \multicolumn{1}{c}{MPJPE $\downarrow$} & \multicolumn{1}{c}{PA-MPJPE $\downarrow$} & \multicolumn{1}{c}{N-MPJPE $\downarrow$} & \multicolumn{1}{c}{PVE $\downarrow$} & \multicolumn{1}{c}{PA-PVE $\downarrow$} & \multicolumn{1}{c}{N-PVE $\downarrow$}   \\
		\midrule
		  ProHMR~\cite{kolotouros2021prohmr} & 95.1 & 59.5 & 93.2 & 109.6 & {74.9} & 108.4 & 126.4 & 70.3 & 111.8 & 152.9 & 86.1 & 126.8  \\
    %\cmidrule(lr){1-1}
        +GMM (DP) & 101.1 & 67.7 & 99.6 & 118.7 & 84.4 & 116.1 & 129.7 & 73.7 & 111.2 & 155.3 & 89.2 & 125.9 \\
        +GMM (OP)* & 103.2 & 66.6 & 101.1 & 118.3 & 82.6 & 117.0 & 130.6 & 73.6 & 115.9 & 152.7 & 88.5 & 130.5\\
		+GMM (OP) & 86.1 & 58.7 & 83.1 & 103.6 & 76.9 & 101.7 & 119.0 & 65.8 & 101.5 & 140.7 & 81.2 & 115.7   \\
        +GMM (\textit{Ours}) & $\bm{79.7}$ & $\bm{53.9}$ & $\bm{78.0}$ & $\bm{95.9}$ & \underline{69.7} & $\bm{94.0}$ & $\bm{108.5}$ & $\bm{63.8}$ & $\bm{90.4}$ & \underline{135.1} & $\bm{79.1}$ & $\bm{105.6}$  \\
        \hdashline
        +VPoser (DP) & 96.2 & 61.7 & 94.7 & 116.8 & 80.8 & 114.8 & 125.3 & 69.1 & 108.2 & 151.4 & 85.5 & 123.6 \\
		+VPoser (OP) & 85.6 & 58.0 & 81.4 & 104.2 & 77.5 & 102.1 & 115.1 & \underline{65.0} & 99.7 & 137.4 & 81.3 & 114.6  \\
        +VPoser (\textit{Ours}) & 84.7 & 57.4 & 83.0 & 103.0 & 75.3 & 101.6 & \underline{110.3} & 65.7 & \underline{93.4} & 137.1 & 81.7 & \underline{109.1}  \\
        \hdashline
        +cNF (DP) & 90.8 & 58.9 & 88.1 & 103.2 & 72.7 & 101.4 & 118.7 & 67.2 & 105.2 & 140.2 & 82.1 & 119.3 \\
        +cNF (OP) & 88.5 & \underline{54.6} & 84.2 & 103.2 & $\bm{69.5}$ & 97.6 & 118.5 & 65.5 & 104.9 & 138.7 & 80.6 & 118.8 \\
        +cNF (\textit{Ours}) & \underline{84.5} & {54.7} & \underline{81.1} & \underline{97.6} & $\bm{69.5}$ & \underline{95.5} & 111.0 & 65.1 & 97.5 & $\bm{132.7}$ & \underline{80.1} & 111.8 \\
		\midrule
		PARE~\cite{kocabas2021pare} & 74.5 & 46.6 & 72.9 & 88.6 & 61.8 & 87.2 & 106.8 & 55.8 & 86.6 & 128.8 & 69.3 & 100.1 \\
        +GMM (DP) &  102.4 & 69.2 & 100.9 & 117.5 & 86.9 & 116.9 & 122.6 & 67.7 & 101.4 & 144.9 & 81.9 & 114.4  \\
        +GMM (OP)* &94.1 & 60.4 & 92.4 & 108.7 & 75.6 & 107.9   & 124.4 & 66.5 & 105.8 & 144.5 & 80.5 & 119.3    \\
		+GMM (OP) & 80.8 & 54.4 & 79.0 & 97.4 & 72.9 & 96.3 & 112.1 & 58.3 & 89.4 & 131.0 & 71.6 & 102.2  \\
		+GMM (\textit{Ours}) & \underline{65.5} & \underline{44.5} & \underline{63.6} & \underline{79.6} & \underline{59.4} & \underline{78.1} & \underline{95.2} & \underline{51.6} & \underline{71.4} & \underline{117.3} & \underline{64.7} & \underline{84.9}   \\
  \hdashline
        +VPoser (DP) & 89.7 & 51.0 & 88.1 & 102.5 & 65.6 & 101.6 & 111.7 & 56.0 & 91.2 & 132.2 & 68.3 & 103.2 \\
		+VPoser (OP)& 73.0 & 45.0 & 69.8 & 87.6 & 59.9 & 84.3 & 104.2 & 52.8 & 83.1 & 121.9 & 65.3 & 95.0 \\
        +VPoser (\textit{Ours}) & $\bm{65.2}$ & $\bm{43.5}$ & $\bm{63.4}$ & $\bm{79.3}$ & $\bm{58.0}$ & $\bm{77.6}$ & $\bm{93.9}$ & $\bm{50.7}$ & $\bm{70.9}$ & $\bm{115.1}$ & $\bm{63.0}$ & $\bm{83.9}$ \\
		\midrule
		HMR+~\cite{pang2022hmr+,kanazawa2018hmr} & 83.0 & {52.1} & 81.5 & 98.1 & {70.8} & 96.1   & 119.3 & 62.4 & 101.6 & 144.6 & 78.5 & 117.3 \\
		+GMM (OP)	& 83.0 & 56.2 & 80.9 & 100.3 & 74.9 & 98.7 & 115.7 & 62.5 & 95.8 & 134.8 & 77.1 & 109.3   \\
		+GMM (\textit{Ours}) & \underline{75.0} & \underline{49.4} & \underline{73.0} & \underline{89.9} & \underline{65.7} & \underline{87.0} & \underline{107.2} & \underline{59.4} & \underline{85.8} & \underline{132.3} & \underline{74.2} & \underline{101.2}  \\
        +GMM (\textit{Ours})$^\dagger$ & $\bm{74.5}$ & $\bm{49.3}$ & $\bm{72.6}$ & $\bm{89.6}$ & $\bm{65.2}$ & $\bm{86.5}$ & $\bm{106.6}$ & $\bm{59.1}$ & $\bm{85.6}$ & $\bm{132.4}$ & $\bm{73.5}$ & $\bm{100.9}$\\
        \hdashline
		+VPoser (OP) & 79.1 & 52.0 & 76.2 & 96.8 & 72.7 & 94.1 & 113.4 & 61.2 & 95.6 & 133.3 & 76.9 & 110.0  \\
        +VPoser (\textit{Ours}) & 78.4 & 52.6 & 76.5 & 95.6 & 71.8 & 93.0 & 110.8 & 62.5 & 90.4 & 135.8 & 77.7 & 106.2    \\
        +VPoser (\textit{Ours})$^\dagger$ & 78.2 & 52.5 & 76.4 & 95.4 & 71.4 & 92.7 & 110.6 & 62.4 & 90.4 & 135.7 & 77.2 & 106.2  \\
        %\midrule
		\bottomrule
	\end{tabular}
 }
 	\caption{Detailed results for 3D human mesh refinement using our 2D displacements, OpenPose keypoints and DensePose predictions. The regressed SMPL parameters of ProHMR~\cite{kolotouros2021prohmr}, PARE~\cite{kocabas2021pare} and HMR+\cite{pang2022hmr+} are refined using SMPLify~\cite{Bogo2016ECCV} with GMM~\cite{Bogo2016ECCV}, VPoser~\cite{pavlakos2019smplx} and a conditional Normalizing Flow (cNF)~\cite{kolotouros2021prohmr} as pose prior. * denotes the default SMPLify implementation of SPIN~\cite{kolotouros2019spin} and $^\dagger$ that we fine-tuned our model on training set predictions of HMR+. The unit of all numbers is mm and the best results are in bold.
    }
  	\label{table:smplify_fitting}
   %\vspace{-1.0em}
\end{table*}

\textbf{Evaluation.}
We use the test splits of 3DPW and the newly released dataset RICH~\cite{huang2022rich} which contains outdoor and indoor video sequences with highly accurate 3D mesh annotations and subjects with varied body shapes\footnote{All datasets were obtained and used only by the authors affiliated with academic institutions.}.
We focus evaluation on challenging in-the-wild scenes and the generalization capability to unseen body shapes and camera angles.
No test subject is seen during training.
We report the mean per joint position error (MPJPE) and its scale normalized \cite{rhodin2018cvpr} (N-MPJPE) and Procrustes aligned (PA-MPJPE) variants.
The equivalent metrics to evaluate the per vertex error are denoted as PVE, N-PVE and PA-PVE.
All metrics are measured in millimeters.

\textbf{Data preprocessing.}
Since only SURREAL provides ground-truth textures, we need to compute the texture maps for the subjects in Human3.6M, 3DPW and RICH.
Note that no texture calibration sequence is available for each subject and that the focus of this work is not on reconstructing high-quality textures.
Therefore, we resort to simply back-projecting the image colors from the reference sequence of a subject to all visible vertices and finally calculate the texture map by taking the median color values.
As seen in Fig.~\ref{fig:texture}, the resulting textures often contain visual artifacts and are blurry, especially in the facial area and around the hands.
This is caused by imperfect 3D mesh annotations and is particularly noticeable for 3DPW. 
Furthermore, the reconstructed textures for the subjects in 3DPW are sometimes incomplete since subjects are not always seen from all sides.
We leave the exploration of more sophisticated texture reconstruction approaches ~\cite{iqbal2022rana,xu2021texformer,alldieck2018video,alldieck19cvpr} to future work.

\subsection{Quantitative Evaluation}
To demonstrate the benefits of using our 2D displacement fields for 3D human mesh refinement, we evaluate SMPLify with three different pose priors on 3DPW and RICH, using the initial SMPL and camera predictions of three different models.
We compare the results against fitting with OpenPose (OP) and DensePose (DP) predictions.
For the GMM~\cite{Bogo2016ECCV} and VPoser~\cite{pavlakos2019smplx} pose priors, we use the publicly available SMPLify implementation of SPIN~\cite{kolotouros2019spin} and initialize the fitting process with predictions from ProHMR~\cite{kolotouros2021prohmr}, HMR+~\cite{pang2022hmr+}, and the state-of-the-art model PARE~\cite{kocabas2021pare}.
However, we noticed that the default implementation consistently leads to very poor results, especially if using OP predictions and if the initialization is already good.
This has also been observed by ~\cite{kolotouros2021prohmr,tiwari2022posendf,joo2020eft,kissos2020eccvw}.
To improve the results, we modify the default implementation by 1) removing the bending and camera depth prior term, 2) fitting in the full image space instead of the cropped and 3) using a focal length approximation of $f = \sqrt{w^2 + h^2}$~\cite{li2022cliff,kissos2020eccvw}, where $w$ and $h$ are the width and height of the full image.
We scale the reprojection loss depending on the size of the person in the image and multiply it by $5.0$, $0.4$, $0.002$ for OP, our 2D displacements and DP respectively.
For fitting with the conditional Normalizing Flow~\cite{tabak2010nf,tabak2013nf,rezende2015nf} (cNF) pose prior of ProHMR~\cite{kolotouros2021prohmr}, we use their publicly available fitting implementation.
We only adjust the weight of the reprojection term to account for having significantly more 2D landmarks.
We multiply the reprojection loss by $0.04$ and $0.002$ for our 2D displacements and DP respectively.
While it is possible to re-evaluate our displacement prediction network after each iteration, we did not find a significant advantage over evaluating it once.

The results are shown in Table~\ref{table:smplify_fitting}.
Our predicted 2D displacements lead to the best fitting results in nearly all metrics and settings.
The gap to OP and DP fitting is especially large when using the GMM pose prior, showing that due to our dense and accurate displacement fields, a complex pose prior such as VPoser is not necessary to constrain the pose space.
While OP fitting with our adjusted SMPLify version significantly improves upon the default implementation, the performance still heavily degrades when using the GMM prior with predictions from PARE.
Thus, fitting to OP keypoints is more sensitive to the initialization and has to rely on strong pose priors.
Since DP fitting with GMM and VPoser, and the default SMPLify implementation always lead to a loss in performance, we do not show the numbers for all settings for the sake of visibility.
We found that while the DP model is generally good at detecting pixels that belong to the person, the predicted correspondences between the pixels and the 3D SMPL surface often lack in accuracy, particularly at the boundary between body parts.
Our displacement fields significantly outperforming the DP predictions for fitting shows the benefit of learning 2D displacements in the image space instead of complex pixel to 3D body surface mappings.
Interestingly, using the strong cNF prior of ProHMR leads to improved results even for DP.
The image-conditioned prior limits the pose space significantly more than the generic priors and can thus better handle noisy 2D landmarks.
However, the prior heavily depends on the estimated conditional pose distribution.
Therefore, it is not suitable to use in combination with stronger models such as PARE, since the fitting converges to solutions of similar accuracy as when initialized with ProHMR.
Nonetheless, fitting with the cNF prior also works best with our 2D displacements.
Due to our dense and accurate 2D displacements, fitting on average works best with the lightweight and simple GMM pose prior.

Although we trained our model only with ProHMR and PARE estimates, it generalizes quite well to predictions of HMR+ as can be seen in Table~\ref{table:smplify_fitting}.
We can further improve the performance by generating HMR+ estimates for our training images and fine-tuning on them.
The results are also shown in Table~\ref{table:smplify_fitting}.

To assess the performance upper bound of our approach, we present the metrics for fitting with ground-truth per-pixel fields in Table~\ref{table:gt_fitting}.
We compare the results to fitting with the 25 ground-truth joints corresponding to the OpenPose skeleton, which we generate from the ground-truth SMPL mesh using the linear regressor and dataset given camera.
Therefore, ground-truth values for occluded joints are used as well.
Despite that our displacement fields only regard visible vertices, they still significantly outperform the 25 OpenPose joints in fitting.
As expected, the performance gap is particularly large for the quality of the refined 3D meshes as shown by the per-vertex error.
Interestingly, we found that the GMM prior consistently outperforms the other two priors when using the 25 GT keypoints.

\begin{table}% [t]\scriptsize
	\centering
 \resizebox{\linewidth}{!}{
	\begin{tabular}{lcccccccccccc}
		\toprule
	       \multicolumn{8}{c}{3DPW} \\
		\cmidrule(lr){2-7}
		Method + SMPLify &
		\multicolumn{1}{c}{MPJPE $\downarrow$} & \multicolumn{1}{c}{PA-MPJPE $\downarrow$} & \multicolumn{1}{c}{N-MPJPE $\downarrow$} & \multicolumn{1}{c}{PVE $\downarrow$} \\
		\midrule
		  ProHMR\cite{kolotouros2021prohmr} & 95.1 & 59.5 & 93.2 & 109.6    \\
		+GMM (OP GT)	& 69.5 & 43.7 & 66.3 & 81.6    \\
        +GMM (Ours GT) 	& $\bm{56.7}$ & $\bm{36.3}$ & $\bm{54.6}$ & $\bm{66.1}$  \\
        \midrule
        PARE\cite{kocabas2021pare} & 74.5 & 46.6 & 72.9 & 88.6    \\
		+GMM (OP GT)	& 50.5 & 33.4 & 47.6 & 62.8 \\
        +GMM (Ours GT) 	&  $\bm{41.0}$ & $\bm{26.5}$ & $\bm{38.2}$ & $\bm{48.5}$  \\
		\bottomrule
	\end{tabular}
 }
 	\caption{Refining ProHMR and PARE estimates using SMPLify with our GT 2D per-pixel displacements and GT OP keypoints.}
  	\label{table:gt_fitting}
   \vspace{-1.0em}
\end{table}

\begin{figure*}
\begin{center}
\includegraphics[width=0.95\linewidth]{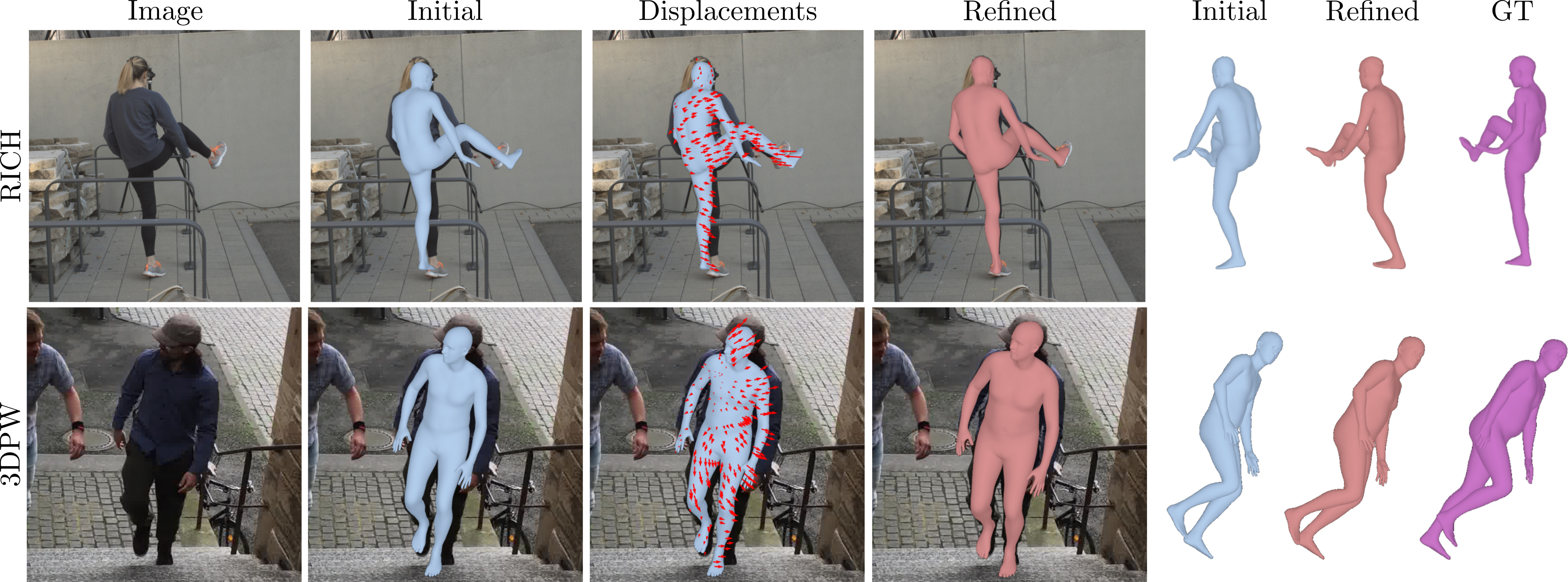}
\caption{Qualitative results on RICH~\cite{huang2022rich} and 3DPW~\cite{marcard2018eccv}. From left to right: input images, initial body estimates, our predicted displacement fields, our refined 3D human models and side views of initial, refined and ground-truth bodies.}
\label{fig:vis_results}
\end{center}
\vspace{-1.0em}
\end{figure*}

\subsection{Qualitative Evaluation}
\begin{figure*}
\begin{center}
\includegraphics[width=0.95\linewidth]{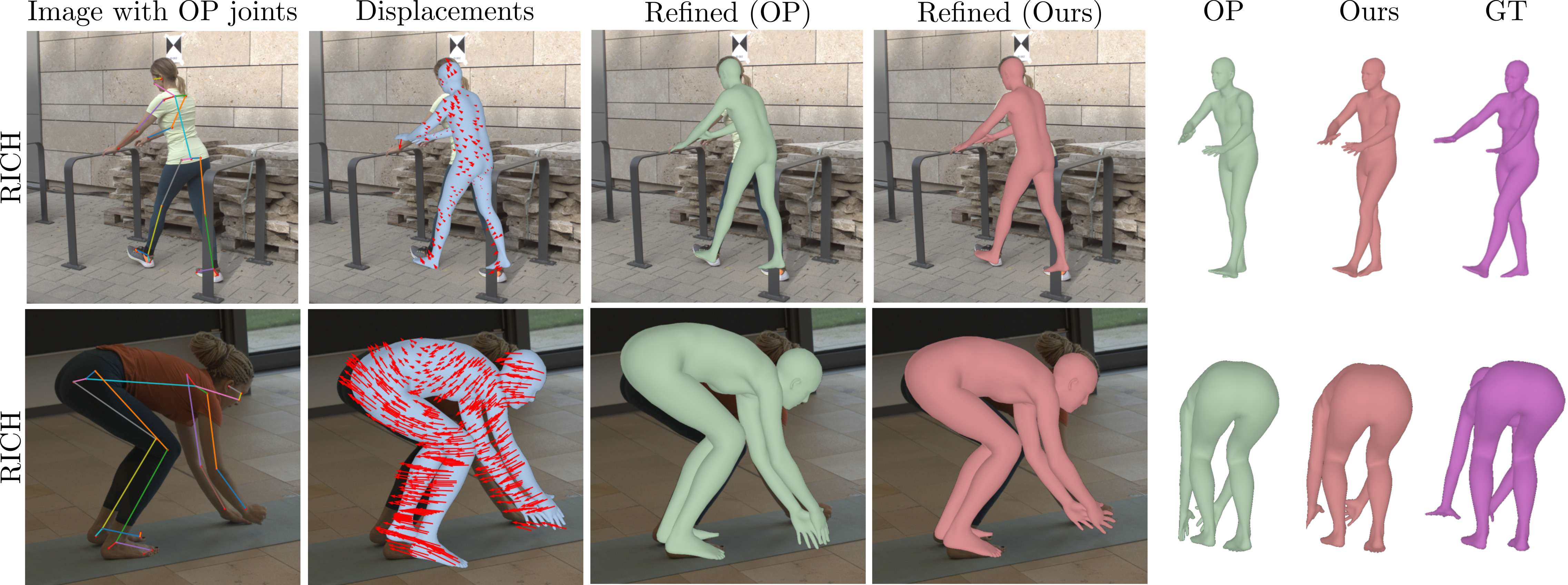}
\caption{Qualitative comparison on RICH~\cite{huang2022rich}. We compare our refined 3D human models (red) with refinements using OpenPose keypoints (green) and the ground-truth bodies (magenta). Best viewed with zoom and in color.}
\label{fig:vis_comparison_op}
\end{center}
\vspace{-1.1em}
\end{figure*}

To better illustrate the degree of improvements, we compare our refined 3D human models with the initial predictions in Fig.~\ref{fig:vis_results} and with refinements using OP keypoints in Fig.~\ref{fig:vis_comparison_op}.
We use the state-of-the-art model PARE and SMPLify with VPoser.
Although the initial estimates are already accurate, our refinement clearly further improves the 3D models.
Compared with the refinements using OP keypoints, we achieve significantly better reconstructions of the spine.
The sparse OP keypoints cannot effectively capture articulation around this area, resulting in incorrect arching of the back.
A larger variety of results and also failure cases can be found in the supplementary material.

\begin{table}% [t]\scriptsize
	\centering
 \resizebox{\linewidth}{!}{
	\begin{tabular}{lcccccccccccc}
        \toprule
		PARE + VPoser &
		\multicolumn{1}{c}{EPE $\downarrow$} & \multicolumn{1}{c}{MPJPE $\downarrow$} & \multicolumn{1}{c}{PA-MPJPE $\downarrow$} & \multicolumn{1}{c}{PVE $\downarrow$} \\
		\midrule
        PARE~\cite{kocabas2021pare} & - & 74.5 & 46.6 & 88.6 \\
        \midrule
        texture + $\mathcal{N}(\bm{0}, \bm{10})$  & 3.98 &  66.8 & 44.0 & 80.6  \\
        texture + $\mathcal{N}(\bm{0}, \bm{30})$ 	&  4.31 & 68.5 & 44.7 & 82.1  \\
        \midrule
        brightness + $\mathcal{N}(0, 25)$ 	& 3.75 & 65.4 & 43.7 & 79.5  \\
        brightness + $\mathcal{N}(0, 50)$ 	& 3.84 & 66.0 & 44.0 & 80.1   \\
        \midrule
        wrong texture & 4.73 & 71.1 & 46.1 & 84.9 \\
        \midrule
		  w/o texture & 4.17 & 66.9 & 44.5 & 80.9    \\
		texture only	& 3.79 & 65.8 & 43.8 & 79.9   \\
        Ours (Full) 	& $\bm{3.70}$ & $\bm{65.2}$ & $\bm{43.5}$ & $\bm{79.3}$    \\
		\bottomrule
	\end{tabular}
 }
 	\caption{Evaluation of the influence of the texture. Results are for refining PARE estimates on 3DPW using SMPLify with VPoser prior. To assess the robustness of our model to noisy and erroneous texture estimates, we manipulate the textures by adding pixel-wise Gaussian noise, changing the brightness and using textures of wrong subjects.}
  \label{table:ablations}
  \vspace{-1.0em}
\end{table}

\subsection{Ablation Studies}
To assess the influence of the texture map, we train a model without texture and one with texture only.
We present the end-point-error (EPE) of the predicted 2D displacements and the metrics after refinement in Table~\ref{table:ablations}.
The appearance information in the textures is successfully leveraged to achieve more accurate displacement predictions.
Additionally providing depth, normal and vertex color renderings further boost the performance.
Despite the importance of the texture, using the model without texture still leads to noticeably improvements.
Thus, if only a single image of a subject is available, it is possible to use the model trained without texture.
We also evaluate the robustness of our model to noisy and erroneous texture estimates, as well as to changes in illumination.
We want to again emphasize that most of the textures are very inaccurate to begin with (see Fig.~\ref{fig:texture}).
For the evaluation, we add pixel-wise Gaussian noise, randomly change the brightness and use the texture of a different subject.
As shown in Table~\ref{table:ablations}, our model is extremely robust to changes in illumination and can also handle pixel noise very well.
The performance most heavily degrades when using the texture of a wrong subject.

\section{Conclusion}
Motivated by the observation that regression-based methods often suffer from coarse alignment between the predicted meshes and image evidences, this work presents an approach to refine initial 3D human mesh estimates using predicted 2D displacement fields.
We learn displacement fields between renderings of the 3D model predictions and the images.
This allows us to exploit the appearance of the persons in form of rough texture maps and additionally leverage 3D information encoded in normal and depth renderings.
Using SMPLify, we demonstrate that dense 2D displacements can be successfully used to improve the image-model alignment and the 3D accuracy of initial 3D model estimates.
Experimental results show that our dense displacements outperform OpenPose and DensePose predictions for 3D human pose and shape refinement.

\textbf{Limitations and future work.} Since our model leverages texture maps, an obvious limitation is that the texture map must be recalculated when the person changes clothes so as to not lose performance.
Exploring an automated way to detect change of clothes and then update the texture could be interesting future research.
Furthermore, as SMPL only captures the undressed shape of the body, extremely loose clothing cannot be modeled.
To improve our approach for loose and complex clothing, future work could employ the SMPL+D model \cite{alldieck2018video}, which extends SMPL by a set of 3D offsets that can be optimized for during the texture calibration sequence.
Finally, we aim to apply our approach to multi-view and motion sequences.
\vspace{-1.0em}
\small{\paragraph{Acknowledgements.}
This work was supported by the Federal Ministry of Education and Research (BMBF), Germany, under the project LeibnizKILabor (grant no.\ 01DD20003) and the AI service center KISSKI (grant no. 01IS22093C), the Center for Digital Innovations (ZDIN) and the Deutsche Forschungsgemeinschaft (DFG) under Germany’s Excellence Strategy within the Cluster of Excellence PhoenixD (EXC 2122).}

{\small
\bibliographystyle{ieee_fullname}
\bibliography{egbib}
}

\newpage
\clearpage
%%%%%%%%%%%%%%%%%%%%%%%%%%%%%%%%%%%%%%%%%%%%%%%%%%%%%%%%%%%%%%%%%%%%%%%%%%%%%%%%%%%%%%%%%%%%%%%%%%%%%%%%%%%%%%%%%%%%%%
% SUPPLEMENTAL:
\renewcommand\thesection{\Alph{section}}
\renewcommand\thesubsection{\thesection.\arabic{subsection}}
\appendix

\section*{Appendix}

%%%%%%%%% BODY TEXT
\section{Implementation Details}

\paragraph{Training details.}
\renewcommand{\thefootnote}{\fnsymbol{footnote}}
We train our model for 600K steps with a batch size of 8 using Adam~\cite{kingma2015adam}.
The learning rate is set to $5\times 10^{-4}$ and we sample training examples from H36M, 3DPW and SURREAL with probabilities $0.4$, $0.3$ and $0.3$.
The images are cropped and resized to $224 \times 224$ while maintaining the aspect ratio.
Additionally, with a probability of $0.5$, we add Gaussian noise to the pose, shape and camera parameters.
We select the PARE prediction with probability of $0.3$ and take a sample from ProHMR with probability of $0.5$.
To also focus on fine-grained displacements, we use ground-truth pose with PARE predicted shape and camera parameters with probability of $0.2$.
Following \cite{kolotouros2019spin}, image data augmentation includes random rotations, scaling and channel-wise pixel noise \cite{kolotouros2019spin}.
Besides, we adopt photometric distortion~\cite{buslaev2020augmentations} and for H36M and SURREAL self-mixing~\cite{chen2021benchmarks}.
The channel-wise pixel noise is also applied on the texture map.

\renewcommand{\thefootnote}{\arabic{footnote}}
\renewcommand{\footnotesize}{\fontsize{6.2pt}{8pt}\selectfont}
\paragraph{Preprocessing details.}
To benchmark our approach, we generate predictions using the latest OpenPose~\cite{cao2019openpose} version (v1.7.0) and a state-of-the-art DensePose~\cite{Guler2018DensePose} model\footnote{\url{https://github.com/facebookresearch/detectron2/blob/main/projects/DensePose/configs/densepose_rcnn_R_101_FPN_DL_s1x.yaml}}. 
For fair comparison, we feed both models with the images cropped around the target subject using the ground-truth bounding boxes.
By transforming the DensePose predictions to points on the SMPL body, they can be used for the reprojection loss \cite{guler2019CVPR}.
For 3DPW, we use the OpenPose detections included in the dataset.
Because RICH only provides SMPL-X bodies, we convert the provided model parameters to SMPL using the official implementation~\cite{smpl_model_transfer}.

\paragraph{Runtime.}
The PyTorch implementation of the displacement field prediction network takes on average 26.4\,ms to process one frame on a RTX4090.
Running our slightly modified SMPLify \cite{Bogo2016ECCV,kolotouros2019spin} implementation for 100 iterations with the reconstructed 2D vertices brings no overhead compared to sparse 2D keypoints and takes around 614\,ms and 769\,ms with the GMM~\cite{Bogo2016ECCV} and VPoser~\cite{pavlakos2019smplx} prior respectively.
Rendering and transforming the per-pixel 2D displacements to per-vertex displacements is in total done in 1\,ms.
For faster evaluation, we run SMPLify in batch mode.
SMPLify with a batch size of 32 takes around 644\,ms and 815\,ms with GMM and VPoser pose prior respectively.
Note that we did not spend any effort optimizing the runtime of our approach.
A highly optimized custom implementation can reduce the fitting time to a few milliseconds~\cite{fan2021revitalizing}, which would enable our approach to run in real-time.
Additionally, by using the refined estimate of the last frame as initialization for the next frame, the 3D pose regressor would only need to be evaluated once.

\section{Additional Results}

\begin{figure*}
\begin{center}
\includegraphics[width=0.95\linewidth]{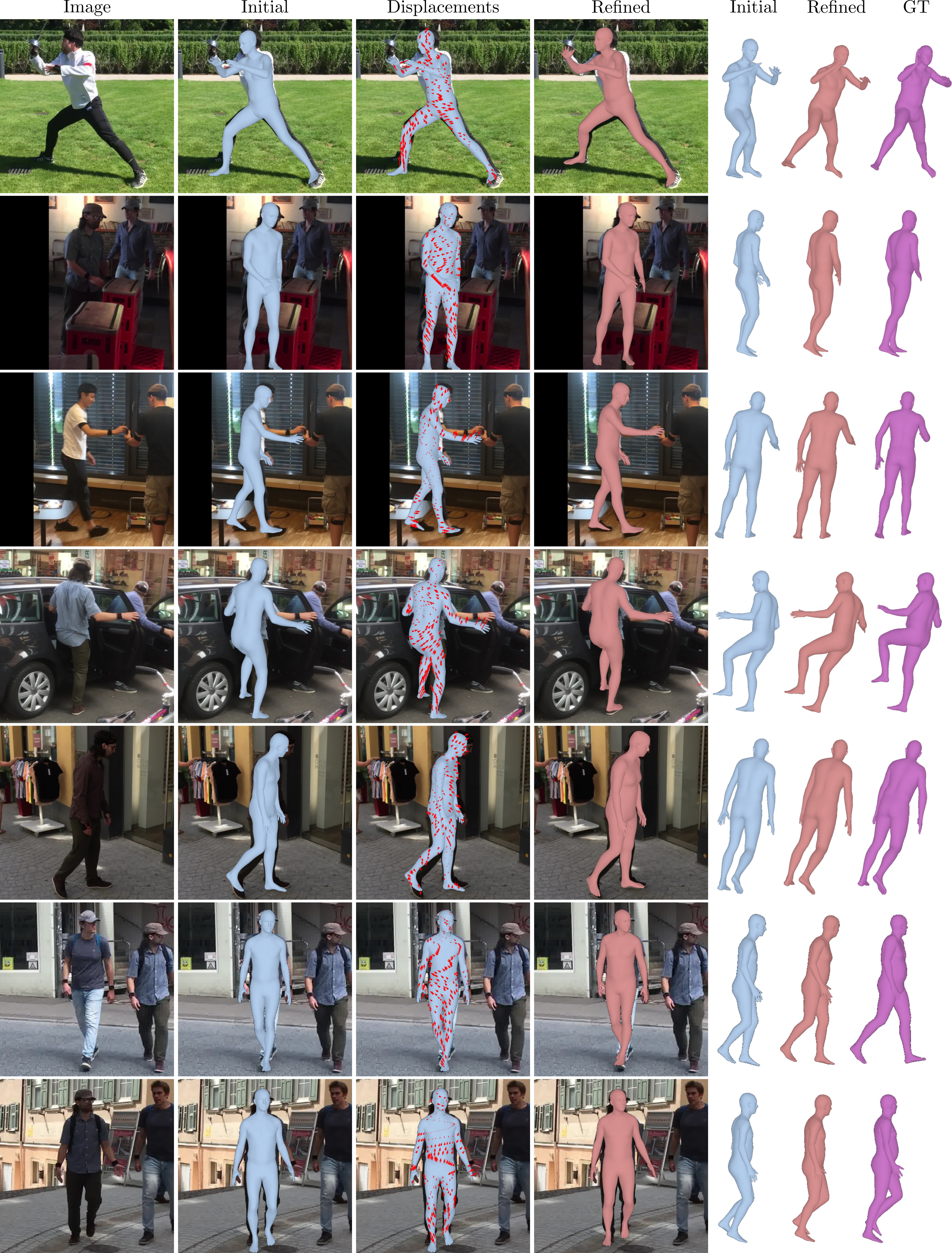}
\caption{Additional results from the 3DPW~\cite{marcard2018eccv} dataset. From left to right: input images, initial body estimates, our predicted displacement fields, our refined 3D human models and side views of initial, refined and ground-truth bodies.}
\label{fig:vis_results_3dpw}
\end{center}
\end{figure*}

\begin{figure*}
\begin{center}
\includegraphics[width=0.95\linewidth]{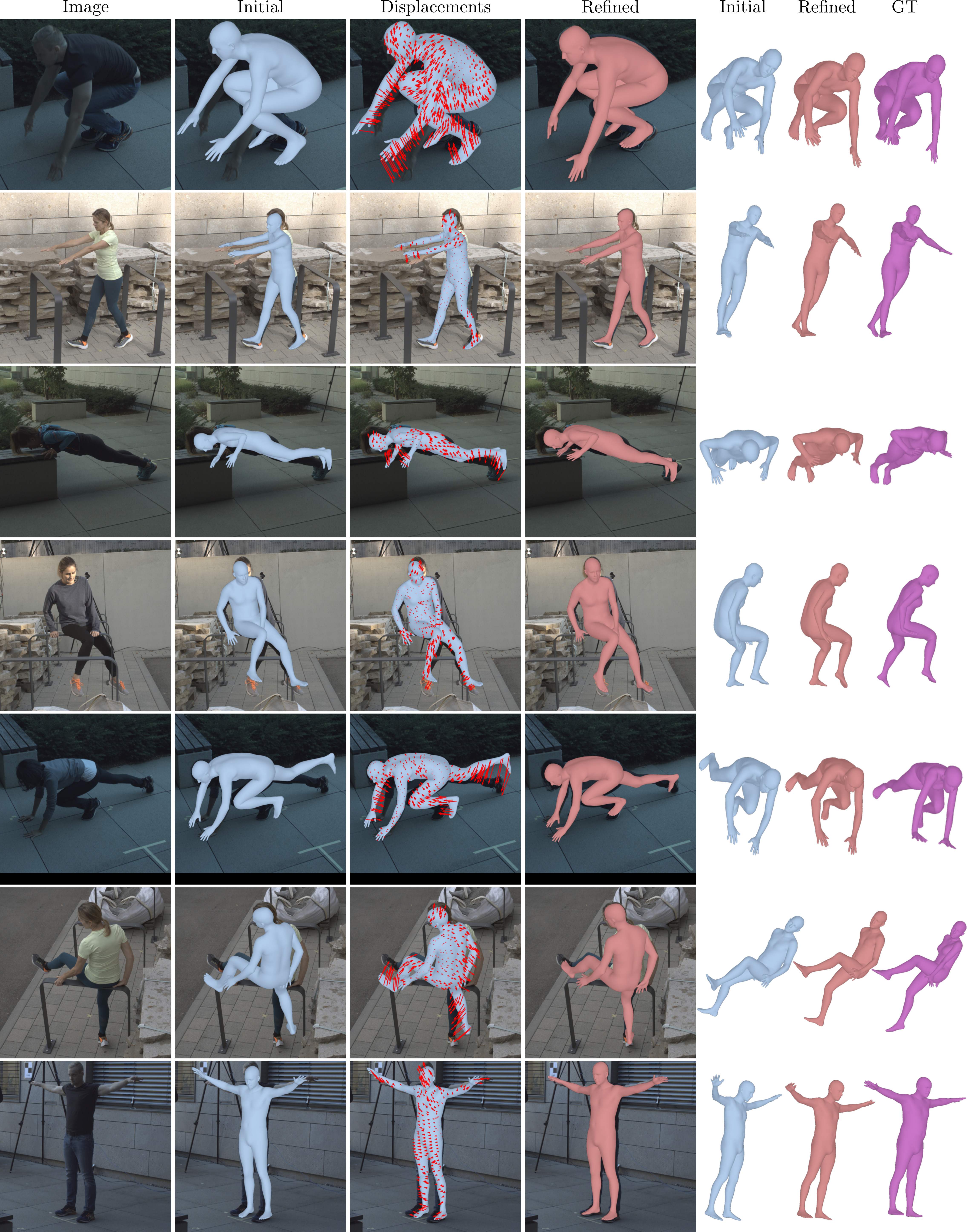}
\caption{Additional results from the RICH~\cite{huang2022rich} dataset. From left to right: input images, initial body estimates, our predicted displacement fields, our refined 3D human models and side views of initial, refined and ground-truth bodies.}
\label{fig:vis_results_rich}
\end{center}
\end{figure*}

We provide more qualitative refinement results on images from 3DPW~\cite{marcard2018eccv} and RICH~\cite{huang2022rich} in Fig.~\ref{fig:vis_results_3dpw} and Fig.~\ref{fig:vis_results_rich}.
We use PARE~\cite{kocabas2021pare} predictions and SMPLify with VPoser.
Our approach generalizes well to different scenes and subjects with varied body shapes, can handle poor lighting and challenging poses, and can even improve fine details such as head rotation.

\begin{figure*}
\begin{center}
\includegraphics[width=0.95\linewidth]{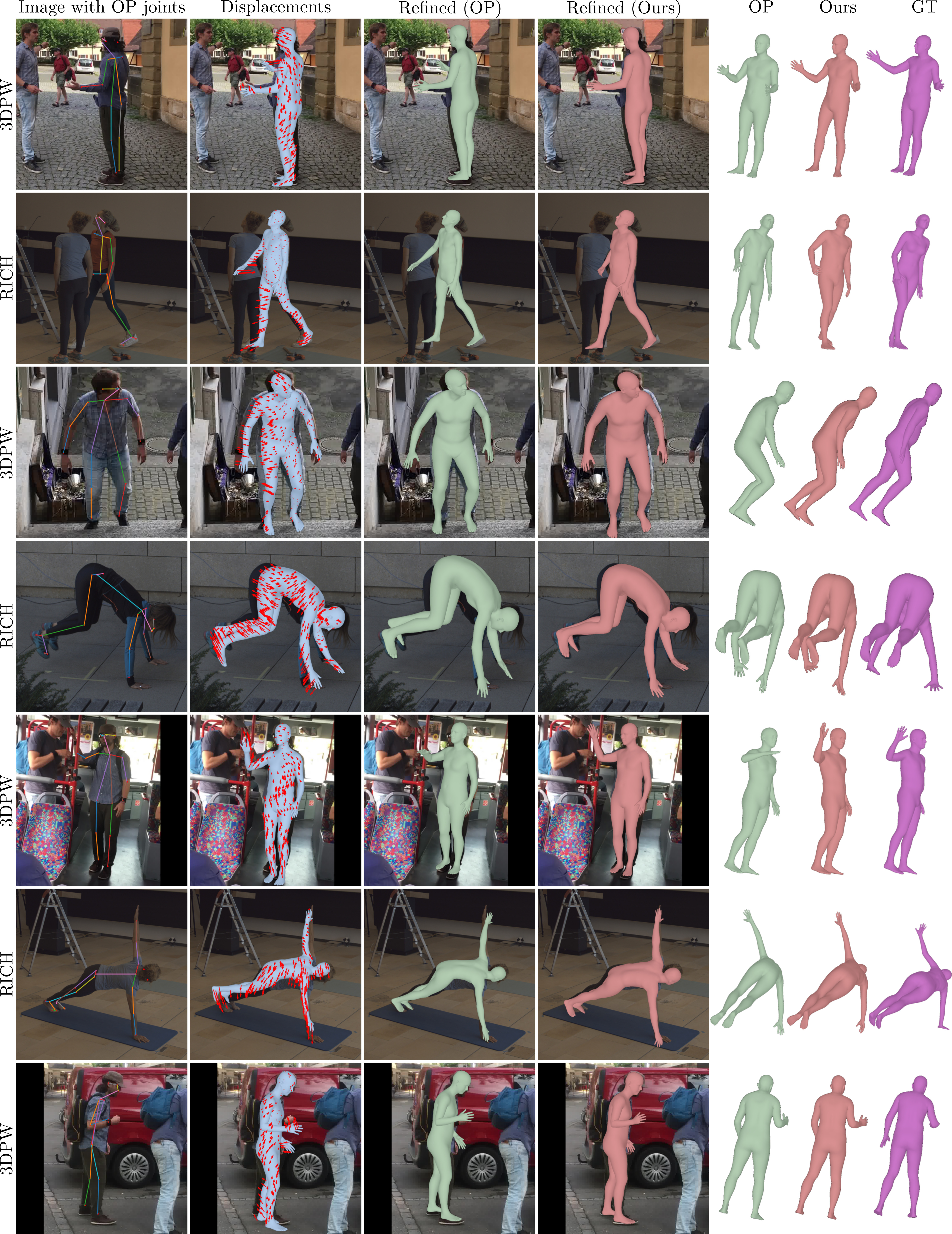}
\caption{Comparison with refinements using OpenPose~\cite{cao2019openpose} keypoints on images from 3DPW~\cite{marcard2018eccv} and RICH~\cite{huang2022rich}.}
\label{fig:op_comparisons}
\end{center}
\end{figure*}

\paragraph{OpenPose comparison.}
We show additional visual comparisons with refinements using OpenPose keypoints in Fig.~\ref{fig:op_comparisons}.
Our approach better refines the reconstruction of the back (row 4, 6), better detects barely visible body parts (row 2, 5, 7) and leads to more accurate depth estimates (row 3).
Additionally, body parts that are visible in the initial SMPL prediction but not in the image can be correctly pushed to be occluded in the refinements (row 1) using our approach. 

\begin{figure*}
\begin{center}
\includegraphics[width=0.90\linewidth]{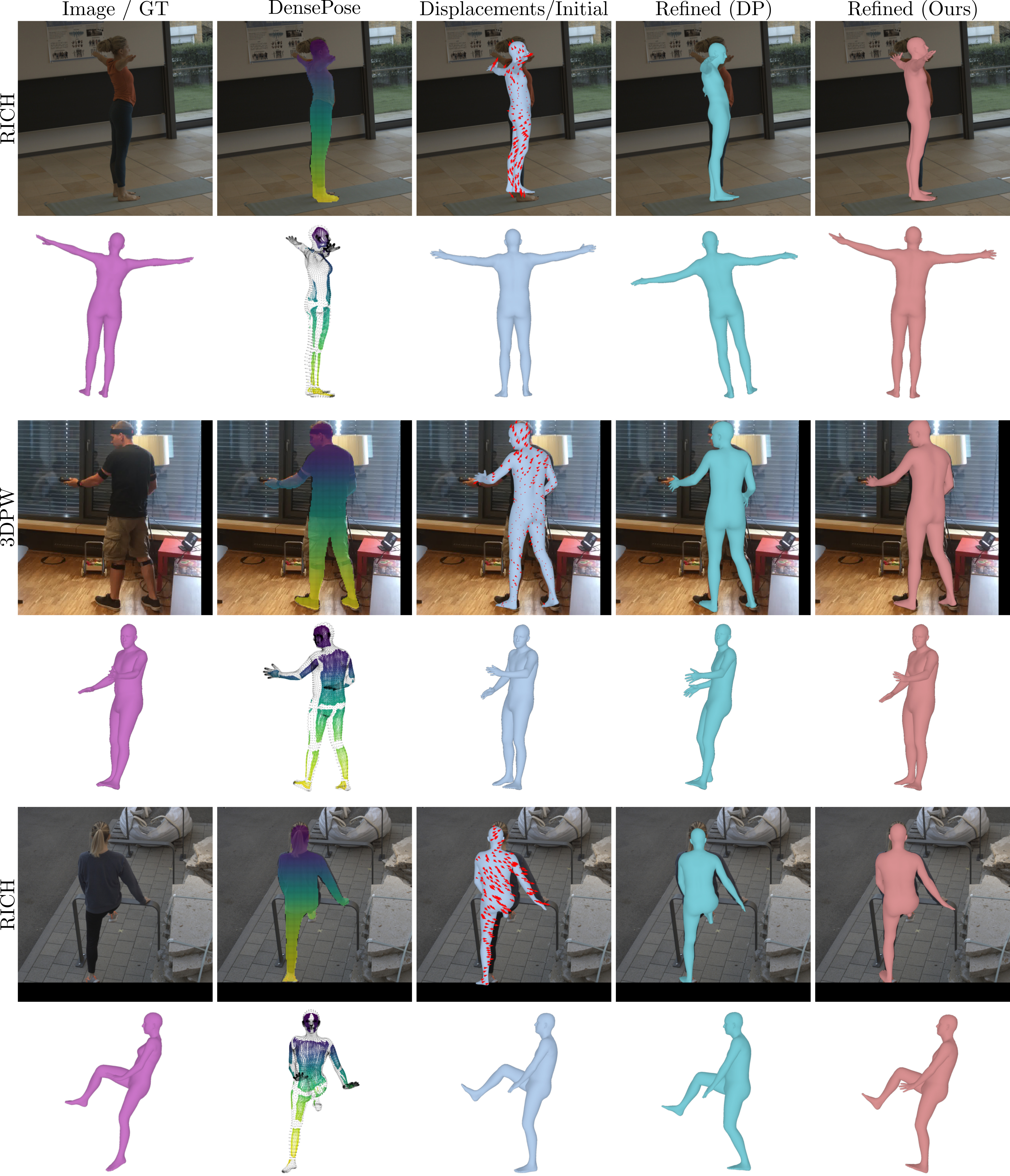}
\caption{Comparing refinements using DensePose~\cite{Guler2018DensePose} on 3DPW~\cite{marcard2018eccv} and RICH~\cite{huang2022rich}. Each person pixel detected by DensePose is colorized in the image and shown on the ground-truth human body.}
\label{fig:dp_comparison}
\end{center}
\end{figure*}

\paragraph{DensePose comparison.}
We visually compare our refined 3D human models with refinements using DensePose predictions in Fig.~\ref{fig:dp_comparison}.
Each person pixel detected by DensePose is colorized in the image and shown on the ground-truth human body.
While DensePose is good at detecting pixels belonging to a person, the predicted correspondences between the pixels and the 3D SMPL surface lack in accuracy.
This is especially noticeable at the boundary between body parts, where no pixels are assigned to even though the regions are visible in the image.
Our approach computes more accurate dense correspondences, leading to significantly better refined 3D bodies.

\begin{figure*}
\begin{center}
\includegraphics[width=0.95\linewidth]{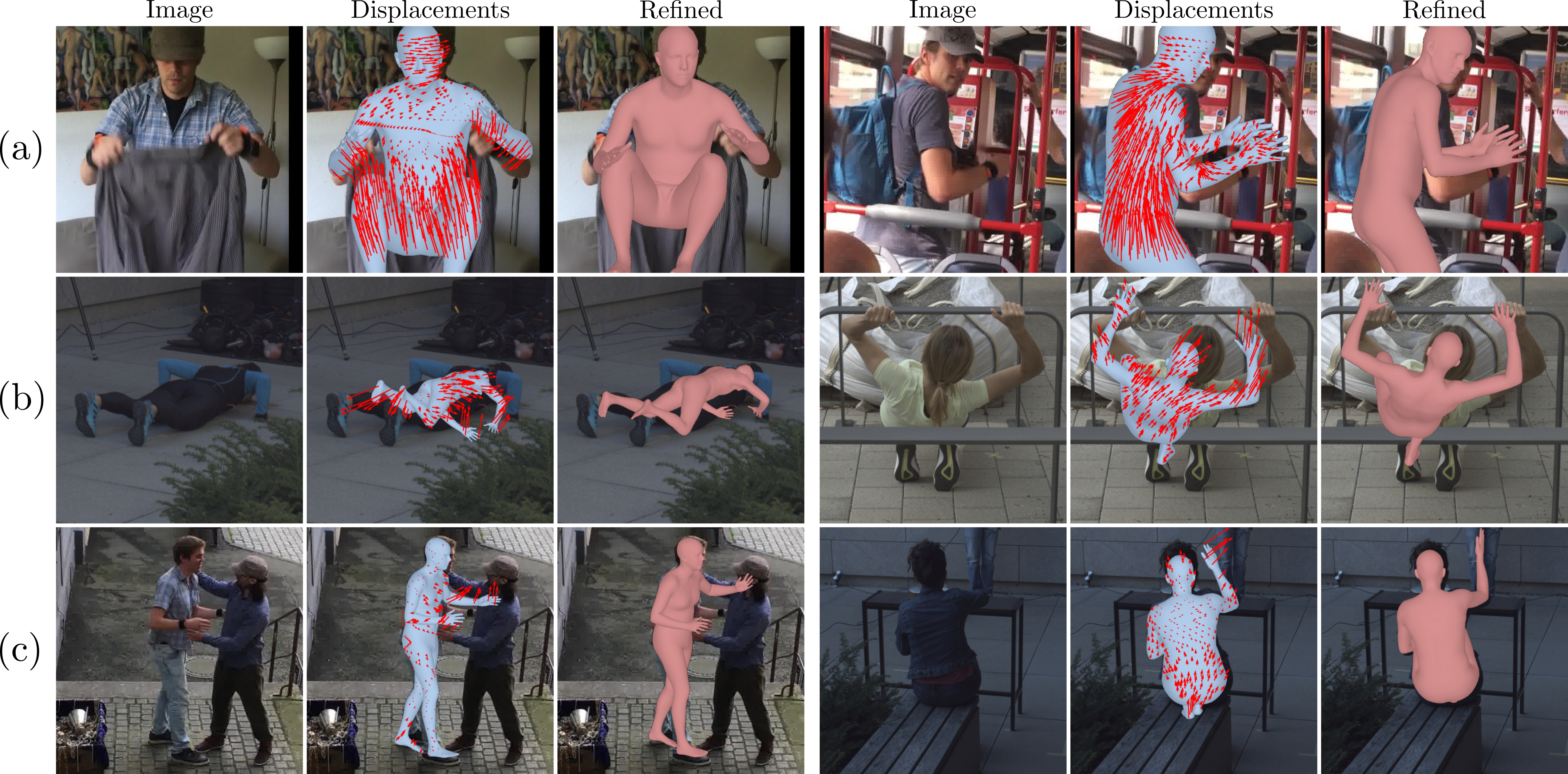}
\caption{Failure cases of our approach with examples from 3DPW~\cite{marcard2018eccv} and RICH~\cite{huang2022rich}. (a) Large occlusions may lead to wrong displacement estimates. (b) If the initial estimate is too far away, displacements may not be enough to fit the model. (c) In some cases of close interactions and overlap with other actors the model may wrongly associate limbs.}
\label{fig:failure_cases_displ}
\end{center}
\end{figure*}

\begin{figure*}
\begin{center}
\includegraphics[width=0.95\linewidth]{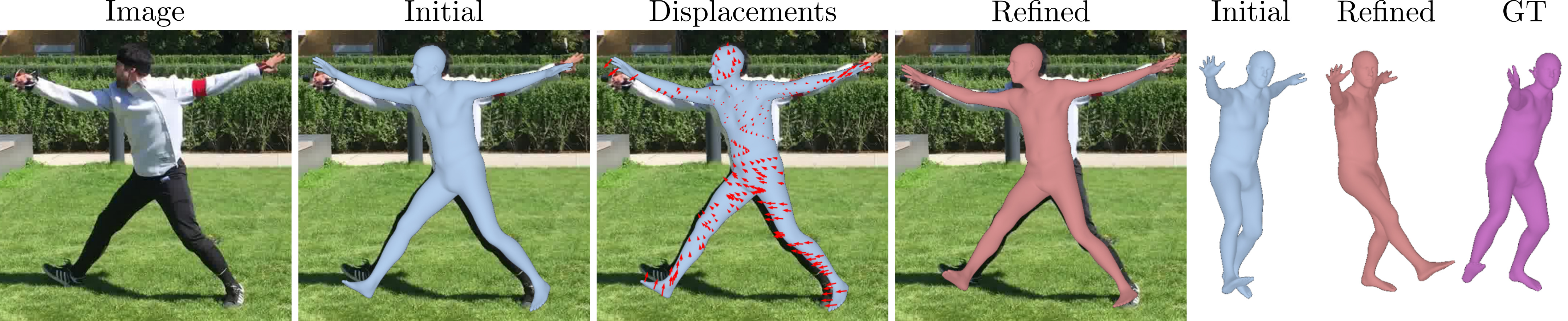}
\caption{In some cases the 2D alignment may be improved by our approach while leading to a worse 3D pose. Example from 3DPW~\cite{marcard2018eccv}.}
\label{fig:failure_cases_depth}
\end{center}
\end{figure*}

\paragraph{Failure cases.}
In Fig.~\ref{fig:failure_cases_displ}, we show a few examples where our network fails to estimate reasonable 2D displacement vectors.
The scenarios range from (a) extreme occlusion, (b) very poor initial body estimates and (c) close interactions and overlap with other subjects.
To improve the performance for large occlusions, it could be helpful to learn visibility masks \cite{Zhang2020CVPR,yao2022visdb} or per-pixel confidence scores.
The problem of wrongly associating limbs could be mitigated by integrating more examples of closely interacting persons in the training set. 
Finally, in some cases our refinement leads to improved image-model alignment but degrades the 3D pose (see Fig.~\ref{fig:failure_cases_depth}).
This is due to the depth ambiguity inherent in monocular 3D motion capture and could be alleviated by regarding multiple images or integrating scene constraints.

\paragraph{Garments with complex texture.}
When evaluating
\begin{wrapfigure}{r}{0.07\textwidth}
\vspace*{-0.48cm}
%\hspace{0.5cm}
\includegraphics[width=0.062\textwidth]{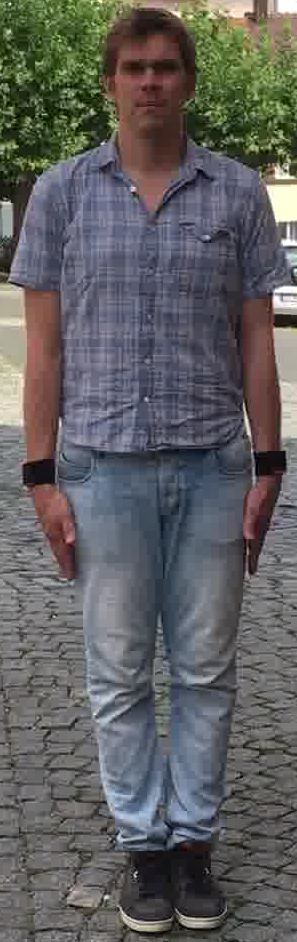}
\vspace{-1.1em}
%\hspace{-3.5em}
\end{wrapfigure}
on the 3DPW test subject with the most complex texture pattern using VPoser and PARE as base model, we achieve an MPJPE of 68.1 and a PVE of 81.2, compared to 75.9 and 90.2 when using OP joints and 75.6 and 88.8 with the base model.
Note that most real world cases allow for a cooperative setting where the person is first turning around in front of a camera, which would allow accurate texture estimation even for complex patterns.

\end{document}